\documentclass[sigconf,nonacm]{cidr-2025} \makeatletter
\def\@ACM@checkaffil{\if@ACM@countrypresent\else
        \ClassWarningNoLine{\@classname}{No country present for an affiliation}\fi
}
\makeatother
\usepackage{graphicx} \usepackage{subfigure}
\usepackage{xspace}
\usepackage{pifont}
\usepackage{tabularx,ragged2e}
\usepackage{makecell}   \usepackage{booktabs}   

\usepackage{enumitem}
\setlist[itemize]{leftmargin=*}
\setlist[enumerate]{leftmargin=*}

\newcolumntype{C}{>{\centering\arraybackslash}X}

\definecolor{darkgreen}{rgb}{0.0, 0.5, 0.0}
\definecolor{darkred}{rgb}{0.5, 0.0, 0.0}

\newcommand{\code}{\texttt}

\newcommand{\myparagraph}[1]{\vspace{1mm}\noindent\textbf{#1}}

\title{Towards an AI Software Factory for Data Systems}

\author[Microsoft, UW-Madison, GitHub]{Anna Pavlenko$^{1}$, Bogdan Crivat$^{1}$, Brandon Haynes$^{1}$, Carlo Curino$^{1}$, Fotis Psallidas$^{1}$, Jaro Slawinski$^{1}$, Johannes Freischuetz$^{1,2}$, Laura Pereira Sanchez$^{1}$, Markus Weimer$^{1}$, Mathieu Demarne$^{1}$, Matthias Jasny$^{1}$, Mauktik Gandhi$^{1}$, Max Bovykin$^{1}$, Mirco Milletari$^{1}$, Purbasha Ghosh$^{1}$, Qiushi Bai$^{1}$, Raghu Ramakrishnan$^{1}$, Rahul Pandita$^{3}$, Sergiy Matusevich$^{1}$, 
Shivaram Venkataraman$^{2}$,
Subru Krishnan$^{1}$, 
Md. Tareq Mahmood$^{1,2}$,
Tiemo Bang$^{1}$, Venkatesh Emani$^{1}$, Xuan Zhao$^{1}$, Yiwen Zhu$^{1}$
}
\affiliation{
     \institution{$^1$ Microsoft, $^2$ University of Wisconsin-Madison, $^3$ GitHub}
  }
\authorsaddresses{}

\setcopyright{none}

\begin{abstract}
AI-assisted coding tools deliver significant acceleration of coding, but only limited impact across the end-to-end software development lifecycle (SDLC)---an Amdahl's law effect!

In this paper, we discuss our progress towards building an {\bf AI SW Factory} that accelerates all the stages of SDLC---Targeting, Coding, Reviewing, and Ops. The AI SW Factory produces a metadata exhaust that enables self-improvement by fine-tuning model weights and updating our World Model (a rich data substrate). 

We focus on {\em Data Systems} and the important class of {\em Evolutionary Coding Tasks} (i.e., those with a measurable objective to hill-climb) and report on 1) scaled deployments at Microsoft (tens of repositories) leading to {\bf 3$\times$ engineering efficiency} above agentic coding and up to {\bf 22$\times$ token efficiency}, and 2) several open challenges.

\end{abstract}

\begin{document}

\maketitle

\section{Introduction} 
\label{sec:intro}
Coding has historically been a costly and complex human endeavor, and as such it has exhibited two defining characteristics. 
First, to ensure correctness and composability of large software systems, the field has developed engineering practices centered around \emph{``decision tracing''}: a rich record of how humans reason about and resolve tasks (e.g., design docs, code versioning, documentation, PR discussions). This has enabled large engineering organizations to collaborate effectively on complex systems and evolve them for decades. 
Second, the inherent complexity of software development motivated the creation of automated tools such as compilers, linters, code style checkers, and testing frameworks, providing fine-grained feedback as early as possible during system development. 

These conditions have proven ideal for AI. The economic importance of coding has driven frontier labs to prioritize it; decision tracing has provided abundant, high-quality training data; and automated feedback has enabled both reinforcement learning for fine-tuning and powerful agentic loops during inference. By mid-2026, agentic AI has made impressive strides in ``solving'' coding.

However, from our vantage point in a large organization, we observe an Amdahl’s Law effect: even a dramatic $10\times$ improvement in coding productivity leads to only modest latency or throughput gains in the end-to-end software development life cycle (SDLC). To improve the whole SDLC, we need to tackle a few challenges:

{\bf Missing Data:} The rich decision tracing that made automating coding possible is either missing or hard to correlate  when looking across other stages of SDLC. 

{\bf Missing Feedback and Verifiability:}
LLMs perform well statistically in bridging the gap between ambiguous natural language and deterministic code. However, they lack formal grounding to support {\bf verifiability/testing} of generated code with respect to (the inherently ambiguous, and possibly incomplete or inconsistent) natural language "specification" or intent. This limits the potential for automated feedback. Things get even harder when we look across other SDLC stages (e.g., choosing what to build, or estimate impact of a feature once deployed at scale). Such high ambiguity creates bottlenecks as human supervision remains critical.

{\bf Harder for Data Systems:} This is further exacerbated for {\em Data Systems}, as they include unique challenges in addition to those shared with all software: (1) system behavior is heavily workload dependent (e.g., table instances, which are separate from code), (2) privacy commitments often prevent direct workload observation, and (3) criticality of the workloads makes any experimental disruption challenging or even unacceptable.

\vspace{-1mm}
\subsection*{AI Software Factory}
\label{sec:ai-sw-factory}

\begin{figure}[pt]
    \centering
     \vspace{-2mm}
        \includegraphics[width=\columnwidth]{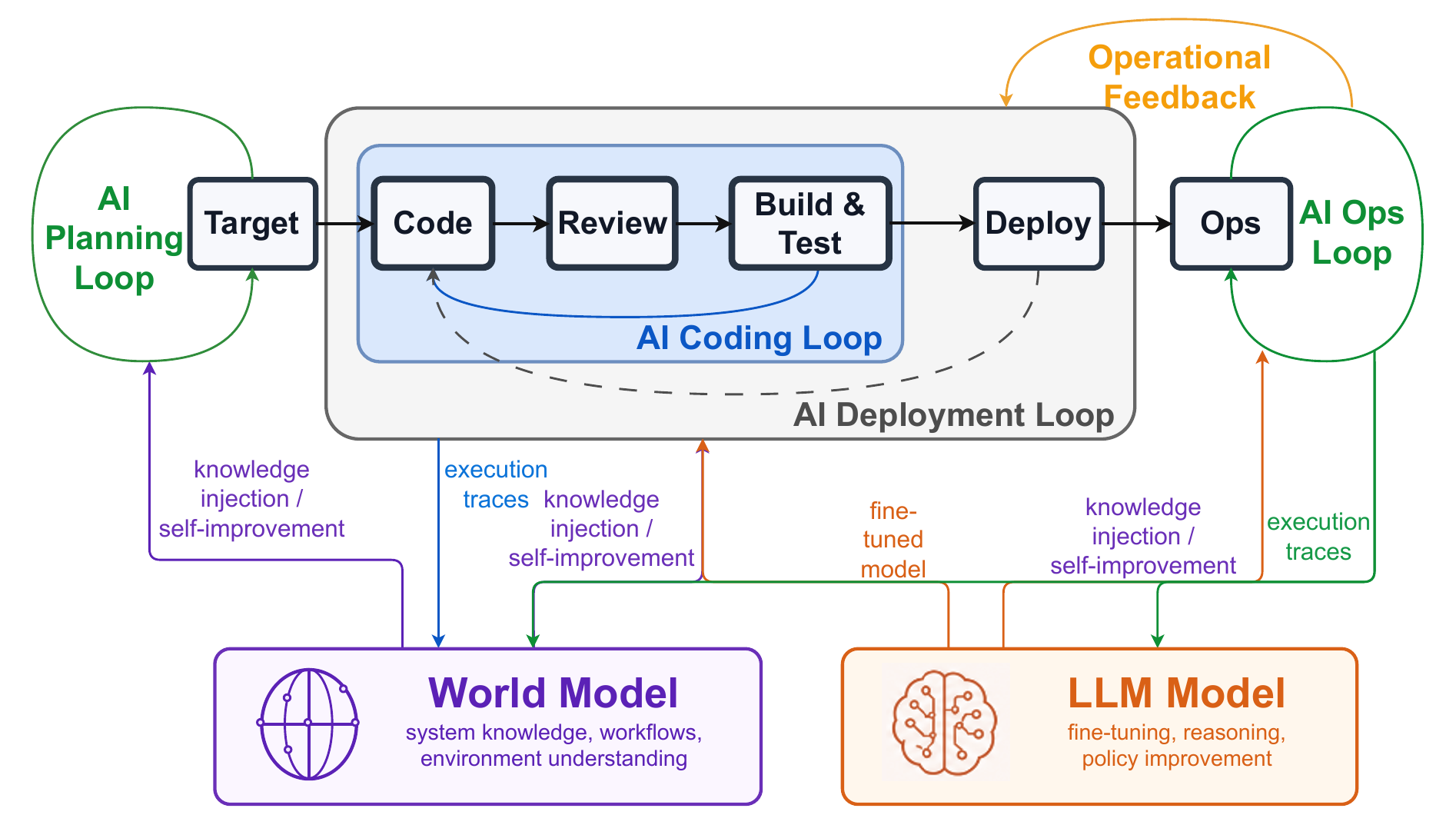}
\vspace{-5mm}
    \caption{A vision for an AI-enhanced software lifecycle. 
}
    \vspace{-5mm}
    \label{fig:ai-vision}
\end{figure}

In this paper, we discuss a long-term vision for an \emph{AI SW Factory} that optimizes the SDLC end-to-end and highlights concrete progress towards addressing these challenges. The general instance of this problem is rather hard, but we show significant progress for narrower but important scenarios. 

{\bf Scoping:}
{\em We focus on {\bf Data Systems}, as they have stable public APIs (e.g., relational DBMS) and clear system semantics (e.g., ACID semantics for transactions), and {\bf Evolutionary Coding Tasks (ECT)}, i.e., coding problems for which we can define an automatically measurable objective (e.g., query latency).} 

By focusing on ECTs for Data Systems we incur less ambiguity: the presence of stable semantics and long-established optimization objectives allow us to automate more of the process by means of increased telemetry gathering and by constructing automated feedback mechanisms. This helps us address the Missing Data and Missing Feedback Challenges.
On the other hand, by focusing on Data Systems we need to manage the increased concerns around workload dependency, mission criticality, and privacy.  Next, we briefly summarize how we automate various stages of SDLC and how to build a learning loop on the entire SDLC (Figure\xspace\ref{fig:ai-vision}).

\textbf{Targeting.} This phase involves determining which new features to build or which system components to optimize. The general problem is layered with ambiguity, but for ECTs on Data Systems we can leverage the right ``data'' to automate good targeting choices. In our experience, this involves two key ingredients: 
\vspace{-1mm}

\begin{itemize}
\item A rich {\em World Model} (\S \ref{sec:world-model}), i.e., a knowledge base capturing the system structure and dependencies, business context, and how these concepts are grounded in observable telemetry. The World Model mediates access to telemetry and profiling, and is continuously evolved by processing incoming evidence (\S \ref{sec:world-model-evoution}).  

\item A {\em Targeting Agent} that given (1) a code repository, (2) a high-level objective, and (3) access to the World Model: ranks potential optimization targets and constructs benchmarks to evaluate any code improvement (leveraging the World Model and its telemetry for an eyes-off production-derived eval). The agent maximizes expected ROI---as lift in target metrics per token spend.  
(\S \ref{sec:agentic-perf}) 
\end{itemize}
\vspace{-1mm}

\textbf{Evolutionary Swarms.} Current coding agents primarily operate in linear or lightly multi-agent workflows, requiring significant human validation after short bursts of AI work. We introduce here a system named {\em Darwin} that employs evolutionary strategies to guide a swarm of agents to optimize software against a well-defined, measurable objective (per the ECT framing), reducing human reliance. Darwin breaks down the overarching goal in a solution space search and efficiently picks which ideas to explore next (\S \ref{sec:darwin}). 

Darwin is inspired by the original genetic+LLM approaches such as \cite{funsearch,novikov2025alphaevolve, gepa, shinkaevolve, openevolve}, but employs an agent-first approach to search that delivers up to 22$\times$ better token efficiency (vs \cite{openevolve}). ECT style solutions have been successfully applied to focused systems research 
\cite{barbarians,letbarbarians}, with this framework we broaden the scope to tackle systematic, large-scale optimization of production systems (\S \ref{sec:prod-evidence}).

\textbf{Reviewing} remains a critical and increasingly costly bottleneck as AI-generated code scales. The evidence-rich approach to targeting and evolutionary coding helps boost confidence, but we argue for the need for systematic {\em Flighting}: i.e., executing new code alongside production and routing a copy of the user traffic to it. This gives us both a correctness check and performance comparison under real workloads. Automating the creation of a privacy-preserving, isolated flighting remains an open challenge (\S \ref{sec:reviewing}). 

\textbf{Ops Automation.} Notwithstanding careful testing/flighting, some changes can only be fully validated at scale. We must therefore complement the impact of a faster paced SDLC by equally faster automated incident management, root-cause-analysis and auto-mitigation. We discuss significant progress in this direction (\S\ref{sec:autosre}).

\myparagraph{Self-Improvement}
Figure~\ref{fig:ai-vision} visually highlights the self-learning loop of our AI SW Factory.  By automating more of the SDLC process we drastically increase the quality and completeness of our ``decision tracing.'' For example, across the Targeting Agent and Darwin we capture detailed telemetry for target decisions, benchmark selections, code changes, performance results, and even consider the impact of code to incidents once deployed in production. 
Importantly, there are multiple ways to ``learn'' from the decision traces. 
For specific facts/skills we evolve the {\em World Model} (\S\ref{sec:world-model-evoution}), while for stable capabilities we fine-tune the model weights (\S\ref{sec:fine-tuning}). This learning loop is a more recent addition to our system, so we have fewer concrete results in this submission, but we expect to have more  by camera-ready / presentation time.

\section{AI SW Factory: Building blocks}
We now discuss in more detail the primary building blocks that comprise our AI Software Factory.

\vspace{-2mm}
\subsection{World Model}
\label{sec:world-model}

The World Model is our primary answer to the \emph{Missing Data} challenge called out  in the introduction. It is the data-management substrate that grounds every step of Figure~\ref{fig:ai-vision}, capturing the evolving shared understanding of workloads, system behavior, contracts, bottlenecks, KPIs, validated hypotheses, and evidence gaps.
It is a bottom-up stack comprising a durable \emph{source of truth} of raw artifacts, a layer of \emph{pipelines and materialized views}, and a set of \emph{viewpoints} through which agents read and write.

\begin{figure}[pt]
    \centering
    \vspace{-4mm}
        \includegraphics[width=0.95\columnwidth]{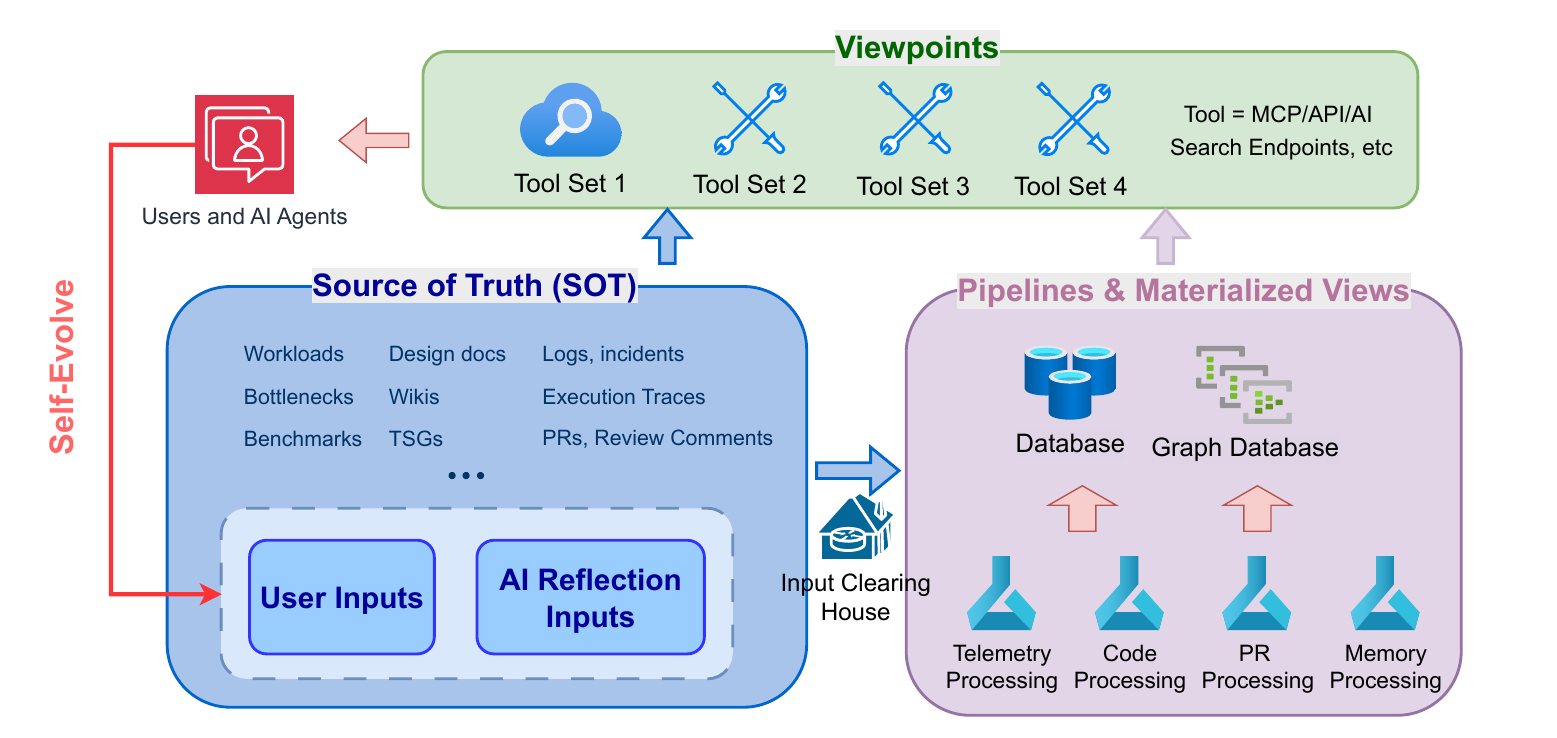}
        \vspace{-4mm}
    \caption{World model architecture.}
    \vspace{-8mm}
    \label{fig:worldmodel}
\end{figure}

\begin{figure}[t]
    \vspace{-4mm}

    \centering
    \includegraphics[width=0.85\columnwidth]{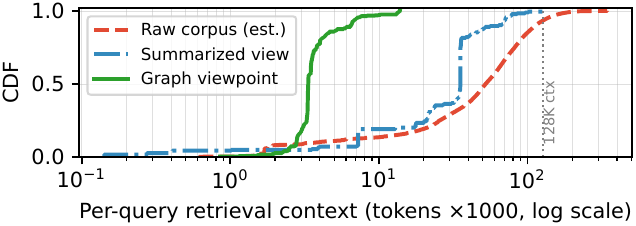}
    \vspace{-3mm}
    \caption{Retrieval context needed for one decision
}
    \label{fig:token_cdf}
     \vspace{-6mm}
\end{figure}

The \textbf{source of truth} is the durable system of record: the collection of raw artifacts everything else is derived from, such as workloads, benchmarks, design docs, wikis, troubleshooting guides (TSGs), logs, incidents, execution traces, and telemetry. For efficiency, pipelines transform this into \textbf{materialized views}.  These views are rebuildable from source, although this can be expensive, complex, and require LLM-based distillation of knowledge. Agents read and write through \textbf{viewpoints}, which are purpose-specific interfaces (i.e., verifiable tools such as MCP servers, APIs, and AI-search endpoints) shown atop Figure~\ref{fig:worldmodel} 
and
layered over the materialized views.  Each viewpoint selects the evidence for a decision, constrains how it is interpreted, and gates which conclusions are readmitted.
Three further properties make this interface load-bearing: \textbf{provenance tracking}, 
which ensures auditability by grounding each action in
a concrete retrieved item that the agent cites;
\textbf{token efficiency}, via a compact, task-scoped context supplied by the viewpoints rather than raw corpora,
which
sharpens focus~\cite{zhu2026graphmindoperationaltracesselfevolving} and yields large savings ($8\times$ on average and up to ${>}100\times$; see Figure~\ref{fig:token_cdf}). Viewpoints also ensure \textbf{access control at retrieval}, limiting information flow to the caller about what is allowed. World Models evolve as evidence accumulates, which we discuss in \S\ref{sec:world-model-evoution}.

\begin{center}
\scalebox{0.9}{
\fbox{\begin{minipage}{1\linewidth}
\textbf{Open Challenge: }\textbf{Generalizing World Model}, we have had good success in creating World Models for specific scenarios in the Data Systems/ECT space, but supporting general purpose world modeling and update is a hard and open problem.
\end{minipage}}
}
\end{center}

\vspace{-2mm}

\subsection{Hotspan: Targeting Agent}
\label{sec:agentic-perf}
{\em Hotspan} is an agent harness that automates an engineer's \emph{targeting} workflow: determining what to change (i.e., a \emph{target}) and how to measure its success. 
Hotspan distinguishes itself from vanilla agents by leveraging provenance and evidence grounded in the World Model to systematically identify potential optimization targets.
To achieve this, its pipeline comprises the following steps:

\myparagraph{Step 1: Priming.} Hotspan first transforms  ambiguous optimization goals (e.g., ``make the service faster'') into actionable objectives. To do so, Hotspan generates true target metrics (e.g., query latency), identifies their constraints (e.g., throughput, cost), and surfaces any missing KPIs. This is a first steering point for a performance engineer to tune the objectives, if needed, and where we start tackling the \emph{Missing Feedback} problem discussed in \S\ref{sec:intro}.

\myparagraph{Step 2: Enumeration.} Next, Hotspan gathers evidence and produces multiple targets. Hotspan queries several information channels via World Model viewpoints (\S\ref{sec:world-model}): (a) {\bf telemetry} surfaces the most expensive regions, propagated appropriately through call graphs; (b) {\bf code analysis} contributes semantic and control-/data-flow queries that elude naive text search, especially in large repositories; and (c) {\bf code review} mines pull requests, issues, and anti-patterns to keep targets consistent with the system's architecture and style. A deduplication step then consolidates targets and aggregates evidence, boosting targets that are corroborated by multiple channels. Channels are extensible and their weights are adjustable. By leveraging multiple channels and correlating evidence across them, we  mitigate the \emph{Missing Data} problem.

\myparagraph{Step 3: ROI prioritization.} Enumerated targets are ranked by expected return on investment, reusing enumeration signals through a cost-and-risk lens. The \emph{return} is a frequency-weighted measure of objective improvement, grounded where possible in counterfactual impact estimates (below), while the \emph{investment} combines three budgets: direct cost (bucketized estimate of token spend and testing/benchmarking cost), stability risk (customers, regions, and call volume on the affected path reported by telemetry, defect propagation through the data-flow graph, and incident history for the affected component), and human comprehension (reviewer effort). Every ROI ratio is weighted by evidence strength, so weakly grounded targets do not outrank well-supported ones. As shipped changes feed their true cost and risk, these outcomes flow back to sharpen later estimates via updates to the World Model.

\myparagraph{Step 4: Packaging.}
Finally, Hotspan packages each selected target as an Evolutionary Coding Task to be optimized by Darwin (\S\ref{sec:darwin}).
To ensure optimizations yield real improvements in production, packaging couples the optimization target with an evaluation harness that provides fast and reliable feedback on correctness and performance.
Packaging is the second point for engineers to steer the direction of agentic coding.

Hotspan currently assembles this harness primarily from test and benchmark suites already present in the codebase. Increasingly, however, we use the counterfactual viewpoint below to evaluate against production-derived workloads.
This approach
embeds optimization targets into their own custom simulators for low-overhead evaluation against production workloads.
The key insight is \emph{scoped fidelity}---executing the bare optimization target and mocking all remaining components that directly interact with it with sufficient accuracy.
We implement this as discrete event simulation for determinism and independence from wall clock time.
The mock implementation is guided by production telemetry to provide realistic inputs to the target and to measure fidelity 
by comparing production and simulation behavior.
We are currently vibe-coding these simulators 
and iterating on the implementation until fidelity is acceptable.
Full simulator generation is becoming feasible, since it is itself an Evolutionary Coding Task, but it hinges on the World Model to provide extensive and reliable information about the targeted system. This counterfactual approach helps us further mitigate the Missing Feedback problem.

\begin{center}
\scalebox{0.9}{
\fbox{\begin{minipage}{1\linewidth}
\textbf{Open Challenges:} {\bf Correctness and Estimating Gains} Automatic benchmark generation and reliably proving that a change is semantically equivalent to the original program and aligned with the user intent, and estimating potential gains with (tighter) performance bounds are open challenges in the general case.
\end{minipage}}
}
\vspace{-1mm}
\end{center}

\subsection{Darwin: Evolutionary Agentic Swarms}
\label{sec:darwin}
Coding agents exhibit a lopsided capability profile. They generate code quickly and cheaply but lack  \emph{architectural taste} and \emph{global self-supervision}.
We sidestep the former limitation by focusing on ECTs in existing (human-architected) systems.
For the latter, existing work 
partially addresses this in two ways.  First, agentic
coding tools (e.g., GitHub Copilot) 
support goal-directed local refinement but lack mechanisms for exploration and optimization across a broader solution space.
Second,
LLM-driven evolutionary search (e.g., OpenEvolve ~\cite{openevolve}) globally supervises individual one-shot LLM code generation but lacks local self-correction. 
Darwin addresses this limitation by synthesizing  
these two lineages: it runs an agentic, memetic variator\footnote{A {\em memetic variator} is a variation operator that produces a candidate which is locally refined before it enters the population.} \emph{inside} an agent-guided evolutionary loop, unifying in-session
self-correction and the global objective search.

Concretely, Darwin comprises a \emph{self-supervising agentic swarm} that
choreographs tens to hundreds of coding agents to efficiently hill-climb a
target objective on a repo-scale codebase. Comparing Darwin to OpenEvolve we observe a $22\times$ token efficiency on classical problems like circle packing.\footnote{\xspace We omit a more complete ablation study we performed due to space limitations.} We argue that this enables us to perform \textbf{Continuous
Optimization (CO)}: a standing, automated counterpart to CI/CD that
continuously drives a codebase toward a measurable objective. Since Darwin
is autonomous, CO can run on off-peak, unused GPU capacity, without direct human supervision. 

In the following subsections, we summarize Darwin's core components and discuss larger scale production evidence in \S\ref{sec:prod-evidence}.

\myparagraph{Memetic variation operators.}
Darwin implements its evolutionary loop 
using a \emph{bounded agent session} (of a powerful coding agent such as Copilot SDK) as its variation operator. 
This is a critical difference from prior LLM-driven evolutionary approaches.
Following the classical notion of \emph{memetic algorithms}~\cite{moscato2003gentle}, each ``mutation'' is a full
code\,--\,test\,--\,debug\,--\,reflect cycle executed inside an isolated
sandbox: the agent clones a repo (or repos), receives a targeted optimization
hypothesis, effectuates a change, runs test/benchmark suites, observes
failures, debugs, iterates, and---critically---\emph{reflects}, emitting
structured hypotheses about what worked and what to try next. Because
agents see compiler errors, test failures, and runtime signals, candidates entering evolutionary selection are already locally debugged, dramatically reducing wasted evaluations.

\myparagraph{Agents steer the search.}
In classical evolutionary computation the variation operator is
\emph{uninformed} and a fixed selection heuristic (e.g., MAP-Elites,
UCB, MCTS~\cite{qd-survey}) is the main source of search intelligence. 
Darwin replaces the fixed
selection heuristic with an \emph{ideator pipeline} of specialized agents 
that continuously reshape and guide the search
landscape.  First, a \emph{seeder agent} periodically generates hypotheses 
for the task by performing a deep-research literature review and synthesizing novel 
hypotheses.  Second, a \emph{refiner agent} continuously monitors the
reflection hypotheses produced by mutator agents and  
distills them into a set of actionable ``future work.''  
Finally, a \emph{matcher agent} jointly scores hypotheses and 
candidate mutations, enabling Darwin to sample jointly
the \emph{hypothesis to evaluate} and the \emph{candidate to mutate} in a single step.
The search direction therefore
\emph{emerges collectively from the agents} and is tastefully curated rather 
than based on a static algorithm.

\myparagraph{Hypotheses as a shared data substrate.} Darwin treats optimization hypotheses as \emph{first-class data objects}. 
Each ``hypothesis'' is a structured,
natural-language record 
with a managed lifecycle: creation (by seeder or mutation agents), weighted
sampling, refinement, matching, and retirement. This gives the swarm a
persistent, cross-generation \emph{memory} that prior systems lack.
Two data-flow
mechanisms make this memory pay off in a massively parallel swarm. 
First, a
\emph{tabu}-style mechanism~\cite{gendreau2003tabu} excludes already-explored hypotheses
from sampling, preventing dozens of concurrent agents from redundantly
re-discovering the same optimization. Second, an exploration-bonus rule adjusts
each hypothesis's effective weight by its historical usage and success
via a contextual-bandit
formulation that balances exploiting proven strategies against exploring
untried ones (akin to agent speculation discussed in \cite{overlords}).
Finally, we observe that hypotheses serve as a stand-in for 
``step size,'' allowing Darwin to adaptively tune the magnitude of each mutation, reduce cost (e.g., using small models where appropriate), and keep sessions within the  model's effective reasoning horizon — which has been shown to be much less than the nominal context length~\cite{hsieh2025rulerv2}.
By combining Hotspan and Darwin we address (for ECTs on Data Systems) the {\em Missing Feedback} challenge described in \S\ref{sec:intro}.

\begin{center}
\scalebox{0.9}{
\fbox{\begin{minipage}{1\linewidth}
\textbf{Open Challenges:}  {\em Cost management} for Darwin in noisy experimental settings (that may mislead search) and for targets with a slow inner loop. {\em Formalizing search} beyond our pragmatic/effective strategy to better understand explore/exploit trade-offs.
\end{minipage}}
}
\end{center}

\subsection{Automated Reviewing}
\label{sec:reviewing}
Reviewing remains a mostly open challenge, though we have made meaningful progress by creating agents that impersonate a subject matter expert based on guiding prompts extracted via the World Model, which, in turn, we learned from historical reviews.
By agentically addressing this feedback prior to human review, we have anecdotally shortened reviewing cycles.  We are currently gathering more systematic evidence.

\begin{center}
\scalebox{0.9}{
\fbox{\begin{minipage}{1\linewidth}
\textbf{Open Challenge:} {\em Flighting} is key to further accelerate reviewing. Building flighting (even for Data Systems) remains a largely manual process, as it is security and privacy impacting core infrastructure. 
\end{minipage}}
}
\end{center}

\subsection{AutoSRE: Automated Ops}
\label{sec:autosre}
No matter how well software is vetted in testing, review, and pre-production environments, it can still cause incidents in production due to scale, diverse workloads, or cloud weather. 

To keep up with the acceleration discussed so far, we must similarly improve the AI-to-human toil ratio for Ops. Our prior AI-aided SRE work~\cite{dricopilot} achieved significant efficiency gains.  Scaling this life cycle demands a far higher degree of automation in detecting, triaging, and resolving incidents.  This is our goal with AutoSRE.

AutoSRE is a concrete instantiation of the World Model of Section~\ref{sec:world-model} specialized to the Ops loop, including self-evolution (\S\ref{sec:world-model-evoution}), and it is how we handle Missing Data and Feedback for Ops itself.

\myparagraph{Incidents, graph, and viewpoint.} The durable system of record for Ops is the stream of production incidents and their attachments: incident-management (ICM) tickets, alerts and firing telemetry, troubleshooting guides (TSGs), mitigation histories, and investigation discussion. Because raw traces are expensive to reason over, AutoSRE distills historical investigations into an incident-investigation graph: an LLM step parses each resolved incident into a small, fixed schema of \emph{problem}, \emph{action}, and \emph{observation} nodes linked by causal and temporal edges, clustering recurring patterns into reusable workflow fragments (as we discuss in ~\cite{zhu2026graphmindoperationaltracesselfevolving}). This graph is a derived, rebuildable, cheaper to traverse {\em materialized view} (\S\ref{sec:world-model}) and the primary {\em viewpoint} agents touch.  Given a live incident, triage and resolution agents enter the graph at nodes matching the symptoms and traverse the highest-evidence paths to retrieve the problem framings, actions, and expected observations that resolved similar incidents in the past, instead of exposing the entire history.

The AutoSRE World Model self-evolves as we discuss in \S\ref{sec:autosre-evolve}.
 
\section{AI SW Factory: Self-Evolution}
The significant increase in automation across SDLC stages presented so far provides us with a non-linear benefit: {\em extensive and well-correlated decision tracing!} Automated agents and tools  meticulously capture every outcome and step of their work, yielding curated data that includes feedback and objective measurement. We now describe how we can ``learn'' from these traces by updating the World Model (\S\ref{sec:world-model-evoution}) and fine-tuning the LLM itself (\S\ref{sec:fine-tuning}). 

\vspace{-2mm}
\subsection{Evolving the World Model}
\label{sec:world-model-evoution}
The World Model is a living knowledge base, evolved by both direct user input and implicit AI reflections. Inspired by continual-learning agent architectures (e.g.,  Voyager~\cite{wang2024voyager}), the AI SW Factory persists and reuses knowledge from prior engineering activities to continually improve.  It does so in the following ways:

\myparagraph{Clearing house.}
Both user and AI reflection inputs collected from user-agent interactions pass through a clearing house (Figure~\ref{fig:worldmodel}), ensuring that the viewpoints  are trustworthy, governed, and compliant. This matters because some of the raw input comes from user-agent interactions (e.g., chat transcripts, logs, screenshots, telemetry) that may carry confidential content, which must be detected, redacted, and tiered before being fed into the shared World Model.  The clearing house leverages the following mechanisms:
\begin{itemize}
    \item \textbf{Tiering with privacy isolation:} context is partitioned by scope, sensitivity, and maturity. Each tier is physically isolated with its own storage, retention, and access policy.
    \item \textbf{Promotion under consent:} personal context is promoted to higher tiers if {\em explicitly consented} and repeatedly observed.
    \item \textbf{Reviewable updates:} applied as WAL-style entries staged through a submitted$\rightarrow$approved/rejected/needs-revision$\rightarrow$processed life-cycle with sequence numbers for time-travel and rollback.
    \item \textbf{Simple revert:} since every materialized view is derived from a source of truth, rolling back an entry just replays the log to the desired point and rebuilds the downstream views.
\end{itemize}
Collectively these mechanisms convert the \emph{execution traces} across the SDLC into durable \emph{self-improvement}: successful trajectories reinforce useful workflows, failures expose stale steps, and reviewed corrections enter shared context, while ensuring end-to-end access control and privacy.  This closes the self-evolving loop of Figure~\ref{fig:ai-vision}.

\myparagraph{Self-evolving AIOps loop.}
\label{sec:autosre-evolve}
As an example, consider the AutoSRE use case of \S\ref{sec:autosre}. Each automated investigation is itself an execution trace (the path taken through the graph, the actions tried, the observations, whether the incident was mitigated). These traces are admitted back through the clearing house and evolve the graph via an ant-colony-optimization (ACO)  mechanism~\cite{zhu2026graphmindoperationaltracesselfevolving}: successful paths deposit ``pheromones'' reinforcing the edges and workflows that worked, unused paths decay, and novel resolutions synthesize new nodes and edges. Over many incidents, the graph automatically reorganizes toward the trajectories that resolve production problems, continually improving AutoSRE.

\begin{center}
\scalebox{0.9}{
\fbox{\begin{minipage}{1\linewidth}
\textbf{Open Challenge:} \emph{Calibrating trust.}
World Model updates are reviewed today; this is safe but costly. Navigating this trade off as we gain confidence in our World Model is an open challenge. 
\end{minipage}}
}
\end{center} 
\subsection{Evolving the LLM: Fine-tuning Loop}
\label{sec:fine-tuning}

The second mechanism updates the model \emph{weights}. Whereas the World Model
(\S\ref{sec:world-model}) captures temporary, revisable facts and skills, the
fine-tuning loop internalizes \emph{stable capabilities}---the kind of judgment
we want the model to exercise by default, cheaply, on every future run. We observe that the improved decision tracing
discussed above turns each Hotspan (\S\ref{sec:agentic-perf}) + Darwin (\S\ref{sec:darwin}) run into a
\emph{free supervised signal}: evolutionary search does not merely
produce a winning artifact, it produces a fully-ordered \emph{lineage} of
candidate edits, each grounded in a natural-language hypothesis and scored by an
automatic, production-derived objective. Our thesis is that \emph{better tracing yields a better model}: the richer and more objectively-grounded the trace, the stronger the learning signal we can distill back into the weights.

\begin{center}
\scalebox{0.9}{
\fbox{\begin{minipage}{1\linewidth}
\textbf{Open Challenge:} {\em Formalize and explore learning}.  What we discuss next is our first take on the problem. We believe there is an interesting open research challenge to effectively leverage the full signal from our long SDLC traces.
\end{minipage}}
}
\end{center}

\myparagraph{From lineage to preferences.}
We turn Darwin lineages into training data without any human labeling. Every
mutation step records a \emph{parent} program and one or more \emph{child}
programs, each of which is executed and scored against the task objective. Two
children and their parent form a naturally-labeled comparison: the higher-scoring edit is \code{chosen}, the lower-scoring sibling is
\code{rejected}, and the shared parent context is the~\code{prompt}. Because both responses are expressed as parent$\rightarrow$child \emph{diffs} rather than whole programs, the same recipe scales uniformly from a single-file kernel to a
multi-file server change, and the model learns to prefer \emph{good edits} over
plausible-but-worse ones; effectively, we frame the problem as a preference learning task. We focus on edits that significantly improve upon their ancestors. 
Crucially, the train/eval split is performed \emph{by run} (and ensuring Darwin has avoided repeated explorations of the same idea), so no lineage leaks across the split and held-out pairs.

\begin{table}[t]
\vspace{-2mm}

\centering
\footnotesize
\setlength{\tabcolsep}{3pt}
\renewcommand{\arraystretch}{1.1}
\scalebox{0.9}{
\begin{tabular}{|l|c|c|l|c|}
\hline
\textbf{Codebase} & \makecell{\textbf{Code Exp.}} & \makecell{\textbf{Darwin Exp.}} & \textbf{Gains} & \makecell{\textbf{Inv.}} \\
\hline
Dataverse & High & None & \makecell[l]{2$\times$ throughput\\4$\times$ mem reduction} & 2--4\,h \\
\hline
Fabric Data Agent & High & Low & 10\% accuracy boost & 3\,h \\
\hline
Fabric Data Warehouse & Low & High & 2.6$\times$ GPU kernel speedup & 4\,h \\
\hline
Xbox & None & High & 2$\times$ code gen speed-up & 16\,h \\
\hline
Copilot CLI & None & High & \makecell[l]{9\% token reduction \\ 12\% reduced cost} & 40\,h \\
\hline
NL2DAX & None & High & \makecell[l]{63\% prompt size reduction \\ 7\% accuracy boost} & 4\,h \\
\hline
Experimental LSM tree & None & Low & 1600$\times$ speedup & 3\,h \\
\hline
\end{tabular}
}
\caption{Darwin Anecdotal Evidence}
\label{tab:case-studies}
\vspace{-9mm}
\end{table}

\myparagraph{Training.}
We fine-tune with offline Direct Preference Optimization (DPO)~\cite{rafailov2023dpo}. We adapt a small open model (Qwen3.6~27B~\cite{qwen3.6}) with a lightweight LoRA adapter~\cite{hu2022lora}. This off-policy strategy approximates the full Reinforcement Learning loop; the natural next step is to use Darwin itself as the rollout generator and reward model, driving the on-policy evolution 
(a l\`{a} AlphaProof~\cite{alphaproof2025}).

\myparagraph{From preferences to behavior.}
We see that the tuned model can \emph{rank} edits; we next ask
whether the learned preference changes how it \emph{acts} as a coding agent
in Darwin. Deploying the fine-tuned LoRA adapter on
\code{circle\_packing}~\cite{novikov2025alphaevolve}, the tuned policy reaches the base policy's best solution
quality with \textbf{72.5\% fewer mutator agents} (11 vs.\ 40) and ultimately attains a
modestly \emph{higher} valid score (sum of radii $2.631$ vs.\ $2.618$, a
$+0.48\%$ gain that closes by 75\% the remaining distance to  
optimal). We attribute this matched-quality speedup to a \emph{trajectory
prior} due to preference training. 

\myparagraph{Why this matters.}
We read these results as early but concrete evidence that end-to-end decision traces unlock real learning: \emph{improving the tracing improves the model}. Scaling this signal from a single high-signal ablation to the full diversity of real-software runs, and to the fully on-policy loop is ongoing work.

\section{Real-World Evidence}
\label{sec:prod-evidence}

Having described our long-term vision, including implemented and future ideas, we now discuss our impact so far in terms of production, OSS, and anecdotal evidence. 

\myparagraph{Anecdotal evidence.}
We have applied Darwin across a broad range of systems. 
On \code{\footnotesize{circle\_packing}}~\cite{novikov2025alphaevolve}, Darwin cost 22$\times$ less than MAP-elites~\cite{qd-survey} and $17\times$ less than~\cite{openevolve}.
Table~\ref{tab:case-studies} reports further evidence of systems we applied it to, the level of codebase expertise by the engineer performing the work, their familiarity with Darwin, gains obtained, and time invested. For example, in Fabric DW we optimize a hash-join GPU kernel. In a matter of hours Darwin derived 40+ software variations, with the best stacking 5 distinct optimizations to deliver a 2.6$\times$ speedup (beyond the production kernel shipping  in \cite{coddspeed}). 
More broadly, we made two surprising observations.  First, Darwin delivered high ROI in nearly every scenario we applied it. 
Second, either tool or codebase expertise alone were sufficient to obtain meaningful results in a matter of hours. The first observation might be biased by our ability to instinctively pick problems that are a good fit for Darwin, so we further validate it over a large set of OSS projects and report on large scale production deployment. 

\begin{figure}[t]
    \centering
\includegraphics[width=0.9\columnwidth,alt={Darwin improvements of OSS projects}]{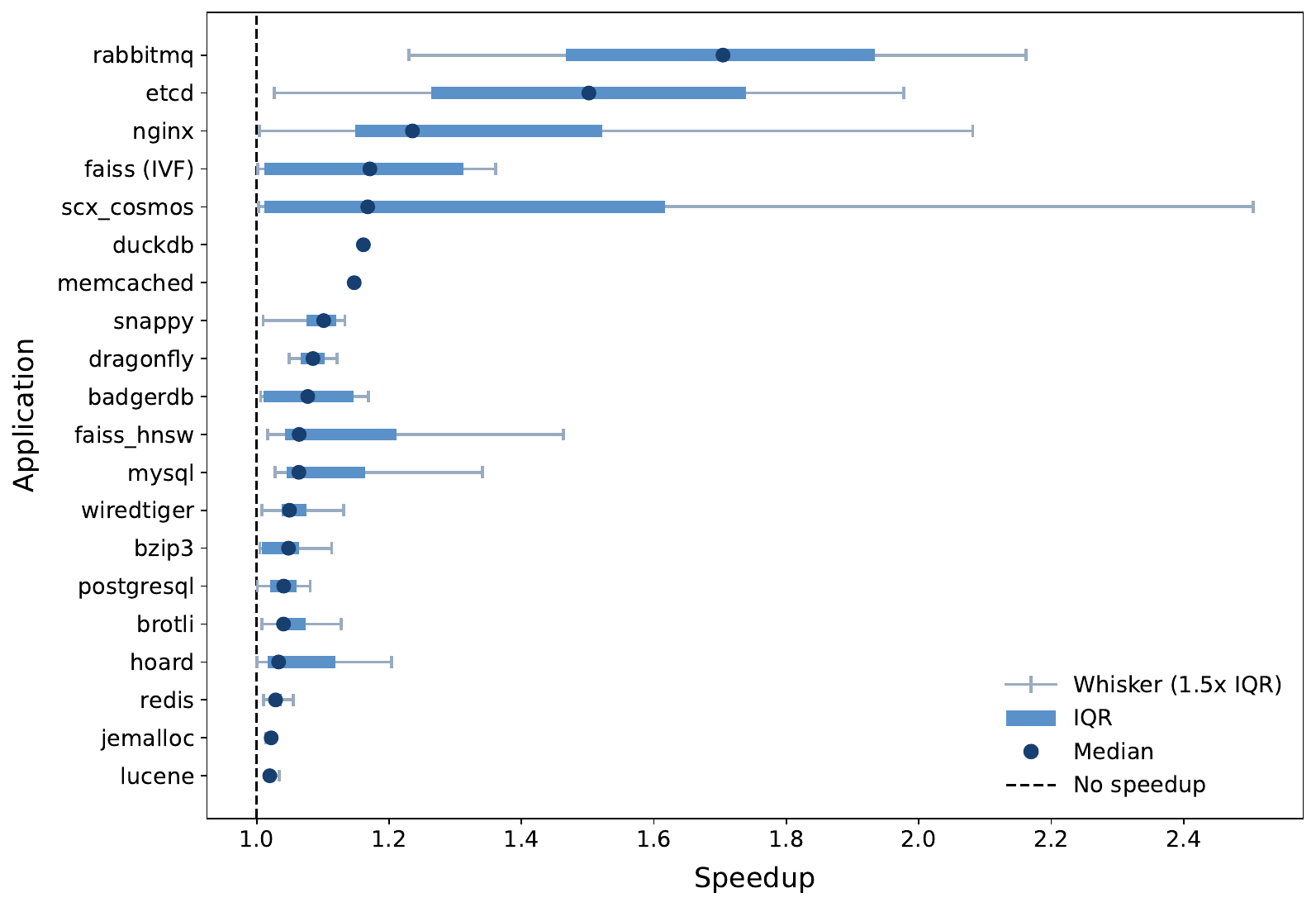}
        \vspace{-4mm}

    \caption{Darwin improvements of OSS projects.}
        \vspace{-2mm}

    \label{fig:darwin-OSS-wins}
\end{figure}

\myparagraph{OSS evidence.}
We apply Darwin to 45 widely-deployed OSS systems. For each, we impose a \$2,500 budget for LLM inference calculated using public API prices. Darwin selects the best-performing mutation, and we report its gains distribution across multiple workloads. We obtained positive (though sometimes modest) improvements across every system we targeted. For the best-performing 20 systems we obtain an average {\bf 13.8\% performance gain} (Figure~\ref{fig:darwin-OSS-wins}).

To gauge whether our roughly {\bf \$100K of tokens} was well spent, we look at the potential impact if these gains were to be conservatively deployed on 10\% of the production instances of the top-20 projects in our set. Estimating impact is inherently hard, so we focus on resource savings (and therefore only on throughput-oriented gains). Using a couple of different methods to estimate OSS project deployment footprint we arrived at estimates of {\bf annual savings in the \$230M--\$616M range}. Even if our estimates are off by 10$\times$--100$\times$, it is easy to justify the small investment. This is strong motivation for \emph{Continuous Optimization (CO)} in data systems. However, even stronger evidence comes from observing the outcome of CO in large-scale production environments, as we discuss next.

\myparagraph{Production evidence.}
Our production toolchain is internally named \textit{Autoperf}, and a dedicated performance team is focused on applying it systematically to repositories across Dynamics / Power platform / Agent 365 (currently tens of repositories with more onboarded daily) with the goal of improving performance and  infrastructure efficiency.  Autoperf leverages a manually-curated World Model (\S\ref{sec:world-model}), a version of Hotspan that uses channels from production telemetry and the Azure Profiler (\S\ref{sec:agentic-perf}), a scale-out deployment of the Darwin evolutionary swarm (\S\ref{sec:darwin}), and an automated adversarial reviewing mechanism to lower the cost of manual reviewing (\S\ref{sec:reviewing}). Autoperf's output is input to engineering review queues. 

\begin{figure}[t]
    \centering
    \vspace{-2mm}
\includegraphics[width=0.95\columnwidth]{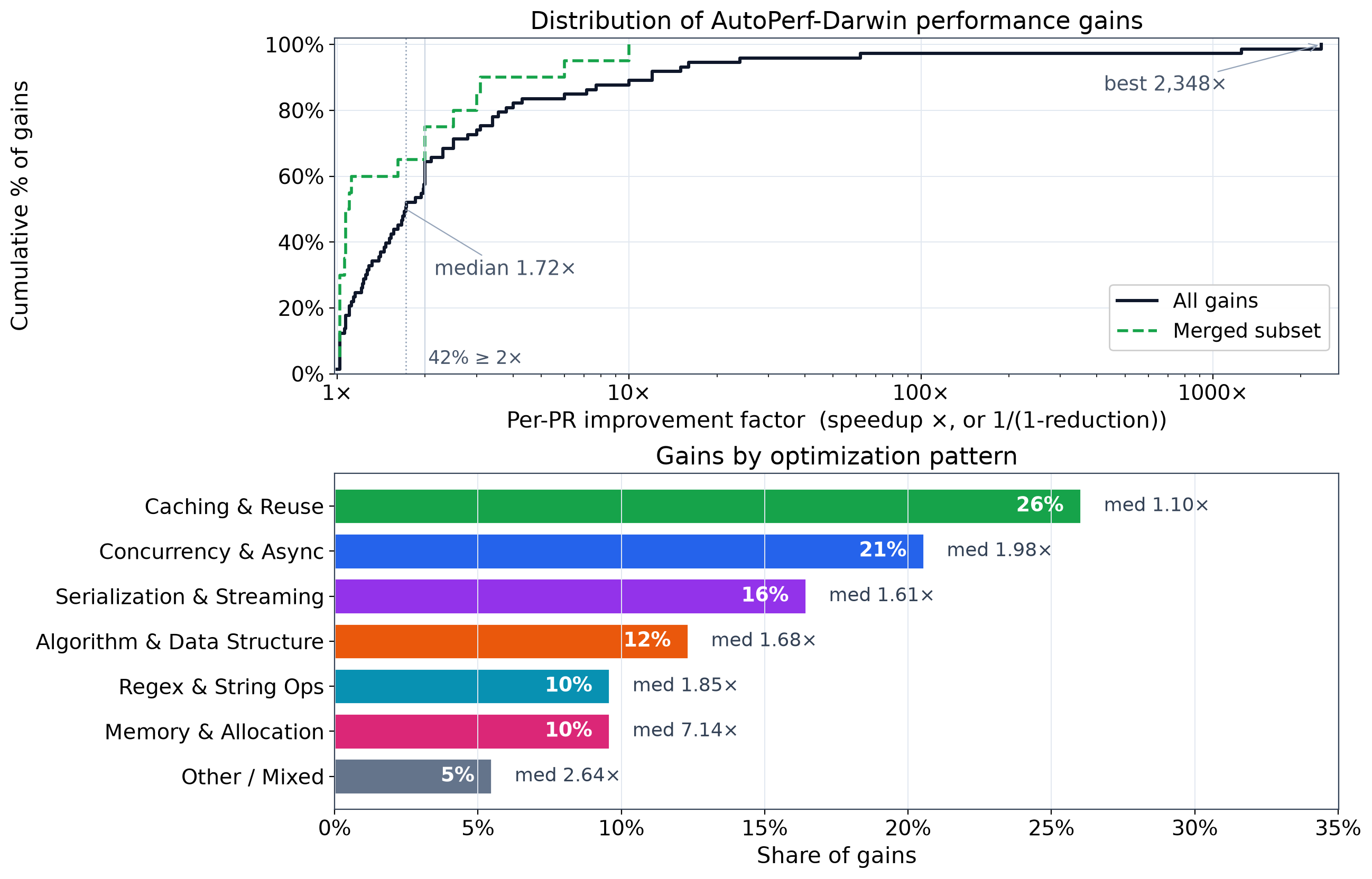}
        \vspace{-4mm}

    \caption{Production Evidence for Autoperf.}

    \label{fig:autoperf-wins}
\end{figure}

Working with the performance team leadership we estimate that about 43\% of the team's time was spent using Autoperf on application scenarios (57\% was in infra building and repo onboarding activities). Focusing on the application work, the team has delivered an astonishing {\bf 0.35 PRs per engineer-hour}. This is estimated to be {\bf 4.7$\times$ more than manual development}, and {\bf 3$\times$ higher than today's best agentic coding tools}.    
Figure \ref{fig:autoperf-wins} shows a CDF of gains from the PRs produced by the team, and a breakdown of what type of code changes led to them. These gains range from latency and throughput to reduction of CPU/memory utilization {\bf (1.72$\times$ median, and 90th \%ile of 10$\times$)}.

\section{Conclusions}
\label{sec:conclusions}
In this paper, we put forth a vision for an AI SW Factory that automates more of the SDLC. We report on an early implementation and its impact both on OSS and Microsoft production-level, large-scale deployments. While many open challenges remain, ECTs for Data Systems have delivered gains far exceeding those of agentic coding alone and easily justified the token cost. This convinces us that we are on the right track to accelerate the SDLC end-to-end and boosts our confidence on the bet of an AI Software Factory!

\balance

{\footnotesize
\bibliographystyle{ACM-Reference-Format-num}
\bibliography{reference}
}

\end{document}